# Adversarial Examples, Uncertainty, and Transfer Testing Robustness in Gaussian Process Hybrid Deep Networks

**John Bradshaw** †‡, **Alexander G. de G. Matthews** † & **Zoubin Ghahramani** †◊
† Department of Engineering, University of Cambridge, Cambridge, UK;
‡ Max Planck Institute for Intelligent Systems, Tübingen, Germany;
◊ Uber AI Labs, USA.

## Abstract

Deep neural networks (DNNs) have excellent representative power and are state of the art classifiers on many tasks. However, they often do not capture their own uncertainties well making them less robust in the real world as they overconfidently extrapolate and do not notice domain shift. Gaussian processes (GPs) with RBF kernels on the other hand have better calibrated uncertainties and do not overconfidently extrapolate far from data in their training set. However, GPs have poor representational power and do not perform as well as DNNs on complex domains. In this paper we show that GP hybrid deep networks, GPDNNs, (GPs on top of DNNs and trained end-to-end) inherit the nice properties of both GPs and DNNs and are much more robust to adversarial examples. When extrapolating to adversarial examples and testing in domain shift settings, GPDNNs frequently output high entropy class probabilities corresponding to essentially "don't know". GPDNNs are therefore promising as deep architectures that know when they don't know.

## 1 Introduction

Deep neural networks (DNNs) have been used to impressive effect in a multitude of tasks such as image classification [33], speech recognition [25] and machine translation [29]. However, these plain models do not capture their own uncertainties [17, 5]. Well-calibrated uncertainties would be very useful in areas such as Bayesian optimization [54] or active learning [53], as well as when using predictions from machine learning algorithms in the real world and noticing when models are in domains away from those upon which they were trained.

Another perhaps arguably related issue of NNs is their vulnerability to adversarial examples. These are images perturbed slightly away from a correctly classified image, taken from the original dataset, for which now the classifier gives very different, incorrect, predictions [57]. This indicates a brittleness in regular DNNs. This can be problematic for classifiers used in security sensitive or safety focused regimes such as malware detection or self-driving cars [12, 6]. There is evidence to suggest that some types of adversarial examples may be able to be prevented by models with good uncertainties [37].

To get better uncertainties from a DNN we can be Bayesian and account for uncertainties in the weights of the network. There exist many techniques for making neural networks more Bayesian [17, 44, 43, 21]. However, some of these methods, and particularly the ones based on sampling, can be difficult to scale to large datasets. This means that Bayesian techniques for NNs are often only used near the top of the network [56, 30] with Kendall et al. [30, §4.1] arguing that the lower layers of a DNN for images may not benefit as much from a Bayesian treatment as they are mainly simple edge and corner detectors.

However, on deciding to make the top of one's deep architecture more Bayesian one does not need to stick with Bayesian linear classifiers. Linear classifiers may be problematic because their softmax function extrapolates high probability predictions far away from a linear decision boundary [17, §1]. Bayesian nonparametrics and in particular Gaussian processes (GPs) [51] would also offer useful probability estimates. Therefore, these have been used at the top of a DNN [26, 9, 61, 63], and have been shown to make good classifiers and regressors.

In this paper we explore using these GP hybrid deep models, GPDNNs. We first describe in the next section the structure of these GPDNN models as well as indicating how they can be trained end-to-end in a scalable manner using variational methods, GPflow [39] and TensorFlow [2]. We then perform three general sets of experiments:

- Classification: here we use these models for image classification and in agreement with the general findings of Wilson et al. [61] find that these models give good accuracy scores even when extended to work on far deeper base networks or in low data regimes.
- Adversarial robustness: we then extend what was previously known about GPDNNs to show that they appear to be more robust to adversarial examples.
- Transfer testing: finally we develop a testing framework for transferring models to new domains and we find evidence that GPDNNs are more robust when used on challenging new domains, as shown by smaller decreases in log likelihood and more uncertain output classes.

## 1.1 Background

Gaussian processes [51] are flexible Bayesian nonparametric models where the similarity between data points is encoded by a kernel function. There are two key problems, however, when working with GPs in complicated domains such as those of images. The first problem is scale; regular GPs are limited to datasets of a few thousand data points. Image datasets, even the small ones, are much larger than this [32, 36]. The second problem is the expressiveness of the kernel, as simple kernels, such as the RBF kernel, cannot build good enough representations to work as effectively with images as DNNs.

In the remainder of this section we will describe the ways people have overcome these problems. We then summarise the structure of the GPDNN model that we will use in the rest of this paper.

**Scaling GPs.** GPs express the distribution over latent variables with respect to the inputs $\boldsymbol{x}$ as a Gaussian distribution, $f_{\boldsymbol{x}} \sim \mathcal{GP}\left(m(\boldsymbol{x}), k(\boldsymbol{x}, \boldsymbol{x}')\right)$, parameterized by a mean function and covariance function. The observed variable, $y$, is then distributed according to a likelihood function, $y|f_{\boldsymbol{x}} \sim h(f_{\boldsymbol{x}})$, given the latent function $f_{\boldsymbol{x}}$. For regression, a Gaussian likelihood is often used for its tractability.

For learning and prediction with GPs one needs the inverse of the covariance matrix. This inversion of a matrix (done usually using a Cholesky decomposition) scales with the number of training data points, $N$, as $\mathcal{O}(N^3)$. This means that in practice regular GPs can only be used on datasets with a maximum of a few thousand data points. To get round this problem inducing points are used. These can be fewer in number than the number of data points and optimised in position [58, 55, 4, 7] or be large in number but placed in a computationally efficient structure [62, 60].

In this work to pick inducing points we use the variational method of Titsias [58]. We use the framework of Hensman et al. [22] that allows these GPs to be fitted via stochastic variational inference, and the extensions of Hensman et al. [23, 24] that allow this to be used with the non-conjugate likelihoods necessary for classification. To summarise this means that we are optimising the following variational lower bound during training [23, §4]:

$$\log p(Y) \geq \sum_{y,\boldsymbol{x} \in Y, \boldsymbol{X}} \mathbb{E}_{q(f_{\boldsymbol{x}})}[\log p(y|f_{\boldsymbol{x}})] - \mathcal{KL}\left(q(f_{\boldsymbol{Z}})||p(f_{\boldsymbol{Z}})\right) \qquad (1)$$

Where $q(f_{\boldsymbol{x}})$ is our variational approximation to the distribution of $f_{\boldsymbol{x}}$ and $\boldsymbol{Z}$ are the locations of our inducing points.

**Kernel expressiveness.** The second issue with regular GPs is the representational power offered by normal kernels. People have built more expressive functions by automatically combining kernels together [14] as well as by composing GPs together [13, 11, 15]. However, even these more sophisticated kernels do not have the representational power to model relationships between complex high dimensional data (such as images) easily.

DNNs on the other hand have been shown to be able to generate good representations, given a large enough dataset, on high dimensional data. Therefore, the idea of putting Gaussian processes on top of DNNs and taking advantage of the good representations that the DNN can learn has been around for a while. Initially, Hinton and Salakhutdinov [26] trained a deep belief network in an unsupervised manner to work out a good feature space for which a GP could be placed on top of. The GP was then trained on top of this network, in a supervised manner, fine-tuning the network's weights. More recently these hybrid models have been built around more regular feedforward DNNs, and again been used for regression tasks [9, 63]. Wilson et al. [61] has also looked at combining this idea with scalable GPs to create a system that can be trained in a stochastic mini-batch manner and be used for classification. These more recent variants of deep kernels have also been used for semi-supervised learning [28].

## 1.2 Structure of our model

The structure of the hybrid GPDNN architectures we consider is compared to more ordinary vanilla convolutional neural network (CNN) structures in Figure 1. We input images to a base CNN and extract features near the top of the network. These are either fed straight into a softmax (architecture A), through a linear classifier with a softmax (architecture B) or into a GP, with the same number of latent functions as the number of classes (architecture C). When feeding the output of the base CNN into architecture B or C we do not need the dimension of this output, $D$, to be equal to the number of classes and so often make it higher. Architectures A and B both represent more ordinary CNNs. We only make distinctions between them in this paper to make the comparison between the number of layers these architectures have compared to the GPDNN clearer and to show when comparing architectures B and C to A that it is not just an additional hidden layer that is improving performance.

When describing the two models in legends and tables we use either NN for architecture A or B and GPDNN for architecture C. We include in brackets after the model name the convolutional network the model is based on, if we are comparing different base convolutional networks.

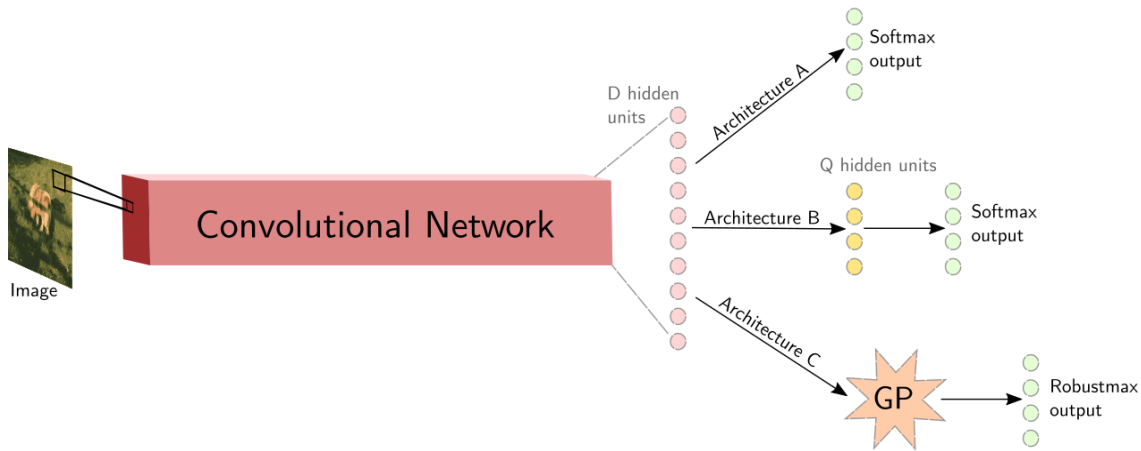

Figure 1: Structure of the models we use in this paper, which are all built on top of the same base CNN structure but with different sized final layers to get D hidden units out. Each architecture learns its own weights for the base CNN. Architectures A and B are regular DNNs with B having an extra fully connected layer. Arch. C is the GPDNN. More explicit details for each individual model can be found in Appendix B.

For the classification from the GP into multiple classes we use a robust max likelihood [24, §4.4], [19]. The robust max assigns probability $1 - \beta$ to that class if its latent variable is higher than all the others and $\beta/(\text{number of classes} - 1)$ if not (see §B.1 for more details).

GPDNNs are written using TensorFlow [2] and the GPflow [39] library for the GP parts. This library is able to back-propagate through all of the GP parts including the Cholesky decomposition [42]. This means we can easily train them end to end.

GPDNNs have a similar structure to the model in Wilson et al. [61]. Our main differences are in the use of a robustmax rather than softmax and the inducing point framework (and implications this holds). As discussed in §1.1 GPDNNs use the variational bound of Titsias [58] as an objective to optimise the inducing points. This means that we are not limited to the additive GP model of Wilson et al. [61] and our GP can operate on the whole hidden layer as input. The tradeoff is that GPDNNs scale worse in inducing point number. Generally, however, we expect that this family of architectures will perform comparably and that the improved results we show on CIFAR-10 in the next section is down to choosing better base CNNs. Likewise, we expect that the new promising properties of these GPDNNs models we show in the following sections may also apply to other hybrid models such as those described in Wilson et al. [61].

## 2 Classification

We start off by looking at how GPDNN models do on classification tasks. In particular we look at the MNIST [36] and CIFAR-10 [32] datasets. We report error rates and log likelihoods as the amount of training data the models have access to is varied.

### 2.1 MNIST

For the MNIST experiments we use as the base CNN the one from the TensorFlow MNIST tutorial[1]. This consists of two convolutional layers followed by two fully connected layers. We refer to this CNN as small CNN (SC) throughout this paper. We assess GPDNNs with linear and RBF kernels.

We take the first 5000 training images and split this off into a validation set. Its size and members remain identical throughout our experiments. We then take the remainder of the training set and train on different proportions of this. We train everything using ADAM [31] for 6000 iterations (batch size of 250). We monitor the validation error and report the test set errors and log likelihoods in Figure 2 for when this validation error is the lowest.

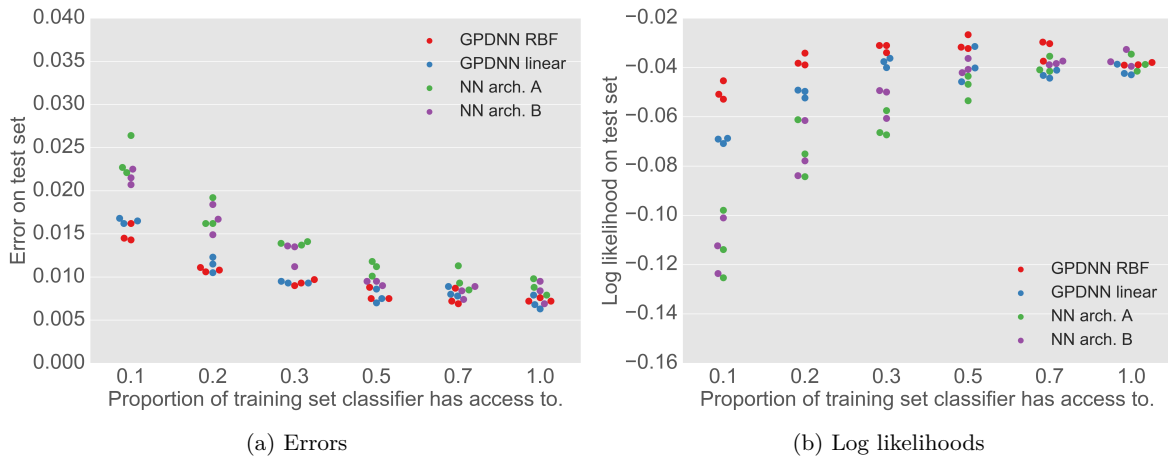

(a) Errors  (b) Log likelihoods

Figure 2: How the hybrid GPDNN compares to the NN model on MNIST test set as the amount of data it trains on is varied. Proportions of 0.1 means we are using 5500 ($0.1 \times 55000$) for training (with a 5000 validation set). A change of 0.001 in error corresponds to 10 images. Please see §B.2.1 for specific details on the architectures.

From Figure 2 one can see that GPDNNs generally have lower error rates and higher log likelihoods, with the gap being most pronounced when one is training on small amounts of data. The architecture B model seems to make some gains with the extra hidden layer, however, it is still short of the hybrid models, showing the extra capacity is not equivalent to one extra layer. Out of the two GPDNN models the RBF kernel does better in log likelihoods compared to the linear kernel, although they perform similarly in terms of error. This could be because the RBF kernel will not overconfidently extrapolate so far in the final layer and so when it is incorrect on points it does not have as high surprise values. We use the RBF kernel with all hybrid models going forwards.

---

[1]available at https://www.tensorflow.org/get_started/mnist/pros

## 2.2 CIFAR-10

We next assess the GPDNN models on CIFAR-10. For the base CNN we use 40 layer deep DenseNets [27]. We were unable to train the hybrid networks starting from random initialisations, so instead we train a regular FC to softmax output for either 225 or 300 epochs before switching to the GP top (effectively we start off training architecture B and switch to C). Further training details are in the appendix.

The errors and log likelihoods are plotted in Figure 3. Effectively we are comparing architecture A for the NN against that of C for the GPDNN. We see that generally the GPDNN performs marginally better than the regular NN in terms of both accuracy and log likelihood. However, the longest training NN has the lowest error when trained on 10% of CIFAR's training set. This may be because it benefits from the larger number of steps, which has overall decreased as the size of an epoch has got smaller.

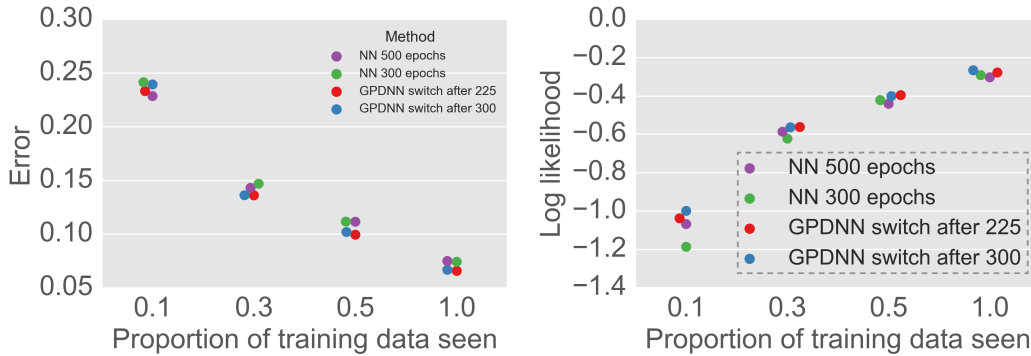

Figure 3: How the GPDNN model compares to the NN model on the CIFAR-10 test set as the amount of data it trains on is varied. The hybrid networks are trained for 500 epochs, with the switching point from architecture B to C shown in the key. The regular NN (arch. A) models are trained for a number of epochs indicated in the key. Note that when training with a smaller proportion of the dataset we run for less steps as the epochs are smaller.

## 3 Robustness to adversarial examples

Szegedy et al. [57] showed that neural networks suffer from adversarial examples. These are examples that are classified incorrectly even though they lie only a short distance (in image space) from correctly classified images. What is worrying about these adversarial examples is that they are often transferable between different architectures [57, 38, 40, 50, 20] and even between different machine learning methods [47, 59]. What is also concerning is that current vulnerable classifiers are also vulnerable to attacks that only operate on a smaller randomly chosen image subspace [16].

One can choose to divide up adversarial attacks into two groups based on whether they are targeted or not. Targeted attacks [57, 48, 10, 34] take the classification model $M_\theta(\boldsymbol{x})$ and find a perturbation $\boldsymbol{\mu}$ such that the new prediction, $M_\theta(\boldsymbol{x} + \boldsymbol{\mu}) = l'$, equals a particular class that they choose. Non targeted attacks [20, 41, 40] try to find a perturbation such that $M_\theta(\boldsymbol{x} + \boldsymbol{\mu}) \neq M_\theta(\boldsymbol{x})$, but do not care what the new class prediction is. In this paper we consider only non targeted attacks.

The attacker wants adversarial attack perturbations to be small in magnitude. This would demonstrate a very fragile and easy to trick model. Whilst ideally you would want some measure based on human sensitivity, such as in Papernot et al. [48], this is complicated to carry out. Therefore, perturbations are more often judged in terms of some distance metric, such as the Euclidean distance from the original image. The attacks are often designed with some metric in mind.

In this section, we want to assess whether GPDNNs have better uncertainties around the images they are trained on and so either maintain correct predictions further out or adjust their confidences appropriately. Adversarial robustness can be assessed through looking at the accuracy/likelihood of perturbed examples and the distance perturbed examples lie for each classifier as well as the predictive entropy of the models on adversaries [3, 49, 37]. We do this for two attacks: the FGSM [20] and the L2 optimisation method of Carlini and Wagner [10]. We mainly demonstrate our attacks on the MNIST dataset, however, Appendix C includes the equivalent attacks on CIFAR-10.

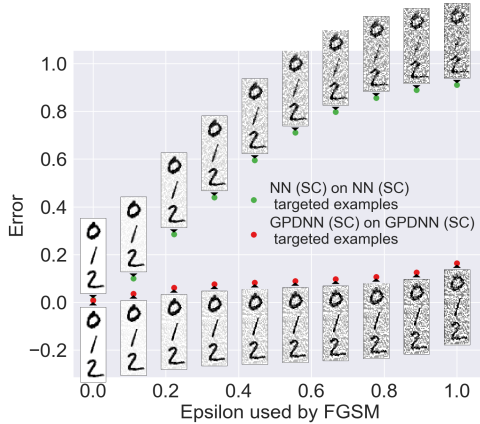
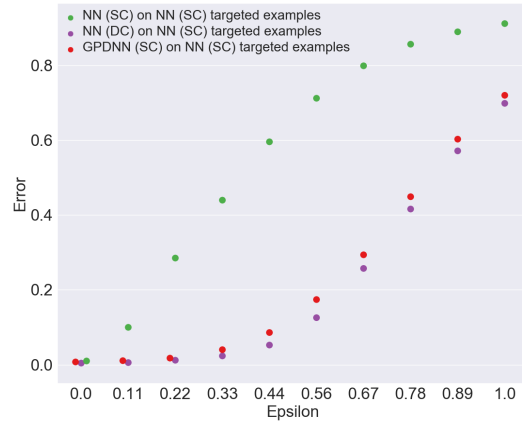

Figure 4: Error from the FGSM attacks (note the dataset is normalised between -1 and 1). GPDNN error increases much more slowly with size of adversarial perturbation.

Figure 5: How the models compare when assessed on the adversarial examples attacking the NN (SC, architecture A) model. See also Figure 6 for LLs and predictive entropies.

## 3.1 The fast gradient sign method

The fast gradient sign method (FGSM) [20] is a very fast method for creating adversarial examples. It is motivated by the idea that neural networks' linearity can explain adversarial examples; to be more specific, a small change in the input image can result in a large change in the feature space that is used by the softmax (if the classifier is linear), which could in turn result in the new image moving across a classification boundary.

In order to find the direction in which to move the input image they add the following perturbation to the input images:

$$\boldsymbol{\mu} = \epsilon \, \text{sign}\left(\nabla_{\boldsymbol{x}} J(\theta, \boldsymbol{x}, y)\right) \qquad (2)$$

Where $J(\theta, \boldsymbol{x}, y)$ corresponds to the model loss with given model parameters, inputs, and outputs respectively. This corresponds to the negative log likelihood for both our GPDNNs and more regular neural network models.

Goodfellow et al. [20, §7] suggest that shallow RBF networks, although not making classifiers that generalise well, are immune to these adversarial examples. They therefore tried using deep RBF models to try to sacrifice some immunity for better generalisation but found finding such models a 'difficult task' when using stochastic gradient descent. As we showed in §2 that the GPDNN deep models we have trained do have competitive error rates, we can now see if they are also more resistant to attack.

To do this we follow Li and Gal [37] by plotting the performance of the classifiers's error rate and entropies on the MNIST test set as the magnitude of the perturbation added is increased. We use the NN (SC) architecture A and architecture C GPDNN classifiers described in §2.1. The attack is implemented using the Cleverhans library [46]. The error as a function of the $\epsilon$ of Equation 2 is shown in Figure 4.

Figure 4 shows the GPDNN is much less susceptible to this attack on MNIST. However, due to the non linear nature of the GP we may expect the directions away at a certain point do not necessarily then push the model back towards the other classes, at least on simple datasets. We therefore also assess the error rates when applying the hybrid architecture on the examples generated on the NN (SC)[2] in Figures 5 and 6. We also here introduce a new model the NN (DD-style CNN (DC)), with the architecture taken from [49, Table I]. This is a deeper CNN of architecture type B: it has four convolutional layers and three fully connected. On normal MNIST it gets an accuracy of 99.6%. We compare against this model to see how the advantages of the GPDNN compare to making more standard improvements in architecture.

---

[2] we use the (SC) notation to indicate that the base CNN for this model is the small CNN introduced in §2. We use arch. A variant of this unless otherwise noted.

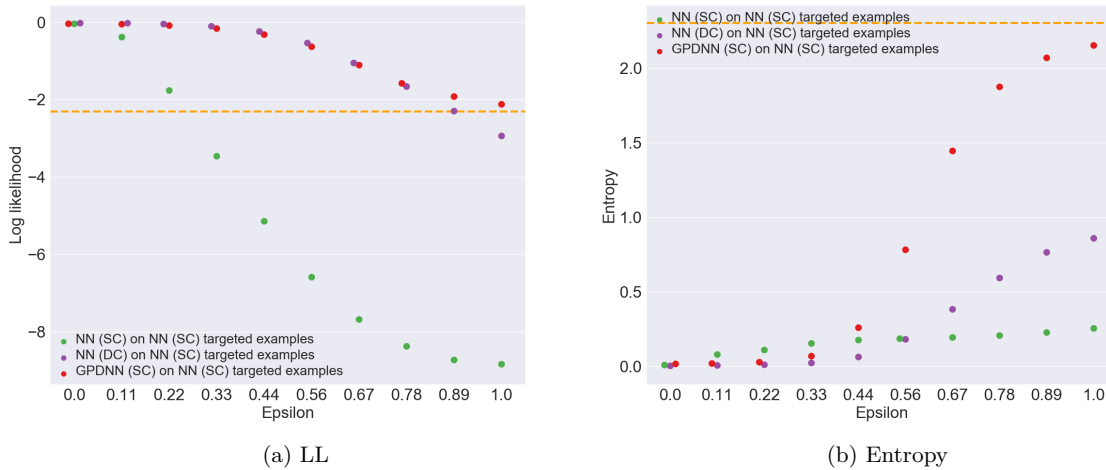

(a) LL

(b) Entropy

Figure 6: How the models compare when assessed on the adversarial examples generated to work on the small CNN architecture A. The DD-style CNN (DC) is the network defined in [49], which is deeper than the others and gets a test error of 0.04% on normal MNIST. The orange dotted lines shows the LL of a uniform class predicting classifier and the entropy of a uniform distribution respectively.

From Figures 5 and 6 one can see that the GPDNN does better in terms of error and log likelihood compared to the NN (SC) model, although worse than when it was performing on its own adversarial examples. On these two metrics it performs comparably to the NN (DC). Importantly, Figure 6b shows that for large perturbations the entropies increase towards the value taken by a uniform distribution. This is an excellent sign as it indicates that the model is not overconfident in unknown regions.

## 3.2 L2 optimization attack of Carlini and Wagner [10]

The attack we consider in this section is the L2 optimization attack of Carlini and Wagner [10]. This considers minimizing the following function:

$$\|\boldsymbol{\mu}\|_2^2 + cf(\boldsymbol{x} + \boldsymbol{\mu}) \qquad (3)$$

Where $c$ is a constant adjusted through line search so that you pick a point where $f(\boldsymbol{x} + \boldsymbol{\mu})$ just becomes negative and $f(\boldsymbol{x'})$ is defined, for the non targeted version, as $f(\boldsymbol{x'}) = Z(\boldsymbol{x'})_l - \max\{Z(\boldsymbol{x'})_i : i \neq l\}$. Here $Z(\boldsymbol{x'})_i$ gives the pre softmax predictions of a model for the class $i$ on image $\boldsymbol{x'}$ and $l$ is the correct class that we wish to move away from. See §C.2.1 for how we adapt this attack to the GPDNN, where we do not have the same pre softmax values available.

We apply this attack to a set of 1000 images from the MNIST test set. We study the differences in the Euclidean distances of the perturbations suggested by this attack on the NN and GPDNN models for each particular image. We notice that 381 of the attacks on the GPDNN appear to fail, this means the attacks optimisation routine does not find an image that fools the network. Of those that succeed it appears that the perturbations are on average 0.529 greater[3] in Euclidean distance. These results suggest that GPDNNs are more robust to these adversarial attacks.

In summary, in this section we wanted to see whether being Bayesian and nonparametric in our top layer would give GPDNNs better uncertainties in the region around points from the dataset and so make them more robust to adversarial attacks. We showed that probably due to its nonlinearity the hybrid GPDNN model is less susceptible to the FGSM attack (at least on simple datasets). When evaluated on the FGSM NN (SC) adversarial examples the GPDNN has much better drop off in error rate and log likelihood, comparable to the deeper NN (DC). Furthermore, its entropies increase far more quickly, highlighting that it realises it has lower certainty in these regions. When analysing the models with the L2 optimisation algorithm the hybrid model appears more robust, with a greater number of attack failures and larger perturbations needed for successful attacks.

---

[3]the images were normalised between -1 and 1 so this distance should be viewed with respect to that, although looking at the examples in §C.2.2 is probably more informative.

## 4 Transfer testing

In this section we want to study how well our hybrid GPDNN models compare to more vanilla architectures in noticing domain shift. When using machine learning in the wild we inevitably will come across data that will not have been sampled from the same distribution as our training data. This is an important problem for instance with self driving cars. A car that has been trained in one environment with one style of road signage needs to understand when it is uncertain in a new environment so that it can take appropriate actions. Being Bayesian in the final layer of our architecture may give us better uncertainties than a normal NN and so be useful in these situations.

Checking one can generalise well has been studied before, for instance see the atypical examples classified correctly in Figure 5 of LeCun et al. [35]. Closely related to this is the idea of domain adaptation where you fine tune models on a new domain [18, 52]. Here we are more interested in the former aspect as well as evidence of the probabilities adjusting in new domains at test time and so do not fine-tune our models.

To evaluate transfer testing we consider how well a selection of classifiers, trained on MNIST, do when used on other image datasets. The image datasets are shown in Figure 7. We convert these to greyscale (where appropriate) and downsample them to the same size as the MNIST images.

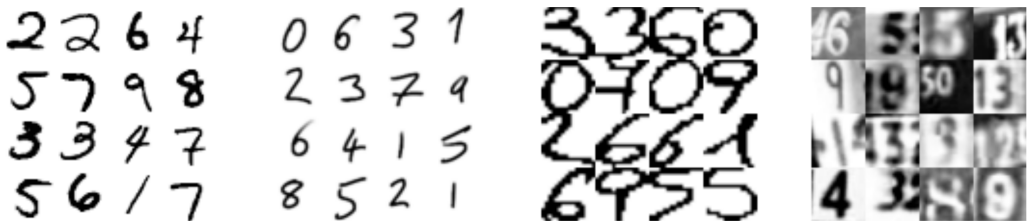

Figure 7: Some random example images from the datasets we test over, from left to right we have MNIST [36], ANOMNIST, Semeion [1, 8], SVHN [45]. Further details are in the appendix. Generalising from MNIST to the three others at test time is clearly an interesting challenge.

We look at NNs and GPDNNs based two base architectures to see how GPDNNs compare to more traditional changes in architecture. To this end we compare the classification performance of the NN (SC, arch A), GPDNN (SC, arch C), NN (DC, arch B) and GPDNN (DC, arch C) models (please see sections B.2.1 and B.2.2 in the appendix for further details of the architectures). The results of this experiment are shown in Table 1. One can see that Semeion and SVHN are much harder datasets. Although the difficulty of SVHN is clear from Figure 1 it is surprising that Semeion is also as hard. It could be because there is no padding in these images with the digits sometimes going right up to the border.

The NN (DC) does better in terms of accuracy. Between the two SC based models there is little difference in terms of accuracy. Interestingly, the GPDNN (DC) performs worse than the GPDNN (SC) in terms of log likelihood on the two harder datasets. What is encouraging is the GPDNN, both for the SC and DC architecture variants, has model log likelihood scores on the two harder test datasets that are substantially better than both regular NNs. Log likelihoods below -2.3 are indicative of classifiers getting overconfident predictions wrong. Both the NN (SC) and NN (DC) suffer from this overconfidence, while the GPDNN gives closer to the ideal *"I don't know"* log likelihood of -2.3.

Table 1: How our three models compare when tested on different digits datasets (all of the models have been trained on MNIST). LL stands for log likelihood. The log likelihood of a uniform predictor on a ten class problem would be at -2.3 (log(0.1)). The appendix provides details and additional experiments testing robustness.

|  | NN (SC) arch A. | | GPDNN (SC) arch C. | | NN (DC) arch B. | | GPDNN (DC) arch C. | |
| --- | --- | --- | --- | --- | --- | --- | --- | --- |
|  | Acc. | LL | Acc. | LL | Acc. | LL | Acc. | LL |
| MNIST | 0.990 | -0.039 | 0.993 | -0.038 | 0.996 | -0.021 | 0.994 | -0.040 |
| ANOMNIST | 0.781 | -1.125 | 0.831 | -0.818 | 0.889 | -0.456 | 0.867 | -0.864 |
| Semeion | 0.249 | -9.609 | 0.320 | -2.841 | 0.450 | -4.836 | 0.317 | -3.458 |
| SVHN | 0.304 | -4.687 | 0.276 | -2.151 | 0.371 | -2.837 | 0.232 | -2.599 |

## 5 Conclusions

In this paper we have explored GPDNN's robustness in classification problems. We have shown that these models use the powerful representations learnt by CNNs to make competitive classifiers. Furthermore, they appear to have taken some of the nice properties of GPs, specifically their better calibrated uncertainties. We show that GPDNNs are less susceptible to the FGSM attack and when shown adversarial examples often have good scaling of predictive entropies with their error rates. We also showed that GPDNNs have better log likelihoods when tested on new domains. This suggests that they do not extrapolate poorly. Overall these are promising results suggesting that such models may be useful in situations where well calibrated uncertainties are important.

### Acknowledgments

We thank Yingzhen Li, Mark Rowland and Alessandro Davide Ialongo for their helpful comments. We wish to acknowledge the EPSRC grants EP/N014162/1 and EP/N510129/1. JB also acknowledges support from a EPSRC studentship.

# A  Why may Gaussian processes offer better uncertainty estimates compared to neural networks?

We argued in the main text that GPDNNs (with RBF kernels) take the nice properties of RBF GPs, namely their lack of overconfidently extrapolating to new areas. We argued that this could be the reason why they were more resistant to adversarial examples and did not show an unreasonable level of overconfidence in new domains.

We further demonstrate this idea in a simple two dimensional example in Figure 8. Here one is building a classifier to split up two half moon shaped classes. Of course in such a simple problem you would not need the power of the complicated representations that the NN is able to learn and an ordinary GP with an RBF kernel performs very well. However, this simple example demonstrates that the GPDNN (RBF) has some of its properties and does not overconfidently extrapolate as far as the models that have linear last layers.

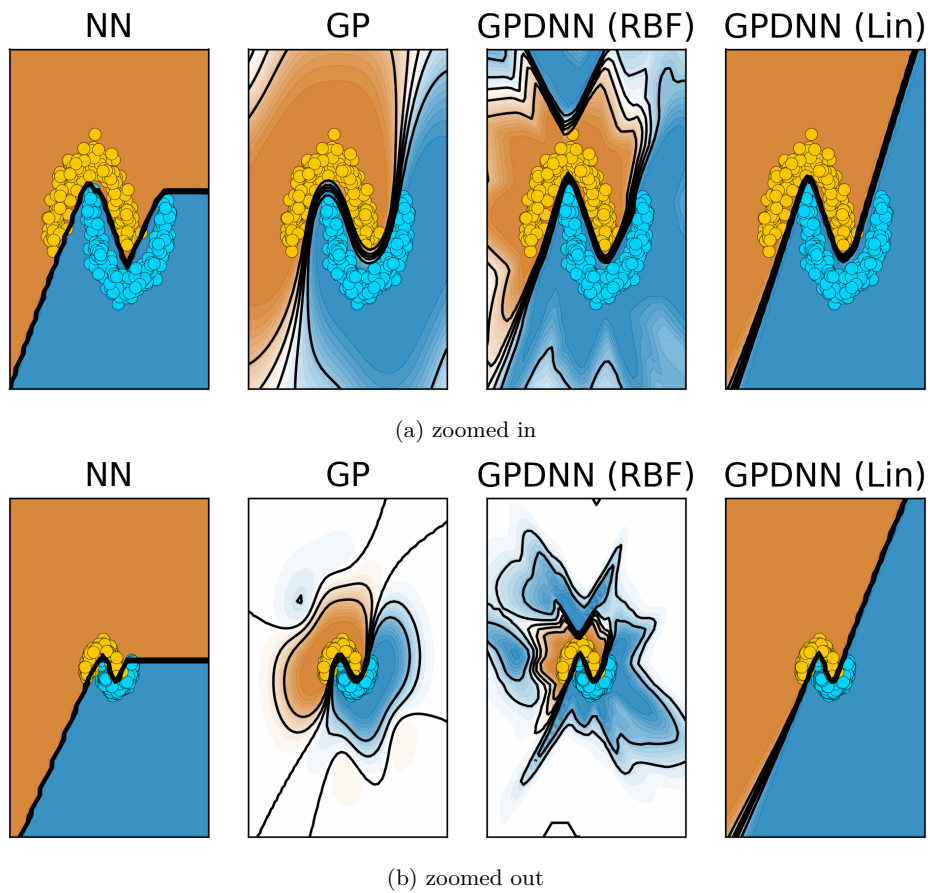

Figure 8: Comparison of the decision boundary of a neural net, a normal GP and a GPDNN on a half moon binary classification problem. The NN and GPDNN models have two hidden layers of 75 units and 10 units respectively with either a linear softmax classifier (for the NN) or a GP (for the GPDNN) on top of this. The regular GP has an RBF kernel and we show a GPDNN with a linear and a RBF kernel.

# B  Model architectures and training details

Here we clarify the architectures we use in the paper. In this section we abbreviate fully connected layers to FC, the white noise kernel to WN and the Linear kernel to Lin. For definitions of these covariance functions see for example Rasmussen and Williams [51, §4], in which the RBF kernel is called the squared exponential kernel. We use rectified linear units as the non-linearities in our networks.

## B.1 The robustmax

The robustmax is defined as [24, 19]:

$$p(y_{\boldsymbol{x}}|\boldsymbol{f}_{\boldsymbol{x}}) = \begin{cases} 1 - \beta, & \text{if } y_{\boldsymbol{x}} = \operatorname{argmax} \boldsymbol{f}_{\boldsymbol{x}} \\ \beta/(\text{number of classes -1}), & \text{otherwise} \end{cases} \quad (4)$$

Note that as we have distributions over the latent variables the gradient is able to flow through these max selecting operations and the likelihood can be evaluated through one-dimensional Gaussian quadrature. We often chose a fixed $\beta$ of $1 \times 10^{-3}$ for our experiments (which are all ten classes), matching what is done in [24, §4.4]. This will ensure that we always assign at least around $1.1 \times 10^{-4}$ probability to each class and so will lower bound our log likelihoods per datapoint at $-9.1$.

For some of the CIFAR-10 experiments in this appendix with the smaller networks we choose to learn the $\beta$ parameter, as these networks often get much lower accuracies than our other experiments and so $1 \times 10^{-3}$ may no longer be an appropriate value. We make clear in the text when this $\beta$ parameter is learnt.

## B.2 MNIST architectures

This section describes the architectures we use on the MNIST dataset in the main paper.

### B.2.1 Small CNN (SC) architectures

For the GP on the GPDNN models we used 100 inducing points.

Table 2: The SC (small CNN) family of models we use on MNIST.

| NN (SC) (arch. A) | NN (SC) arch. B | GPDNN linear | GPDNN RBF |
|---|---|---|---|
| | Conv. (5 by 5, 32 channels) | | |
| | Max pooling (2 by 2, padding is SAME) | | |
| | Conv. (5 by 5, 64 channels) | | |
| | Max pooling (2 by 2, padding is SAME) | | |
| | FC (to 1024 units) | | |
| FC (to 10 units) | | FC (to 100 hidden units) | |
| Softmax | FC (to 10 units) | GP (Lin + WN) | GP (RBF + WN) |
| | Softmax | Robustmax | Robustmax |

### B.2.2 DD-style CNN (DC) architectures

In Table 3 we show the DD-style CNN (DC) architecture. This has dropout rate of 0.5 on the fully connected layer. It is trained for 50 epochs with a batch size of 128 using the ADAM optimiser. For the GPDNN model we use 100 inducing points.

## B.3 CIFAR-10 architectures

### B.3.1 DenseNet based models

The DenseNet [27] model we use is shown in Table 4

A DenseNet block corresponds to 12 convolutional (3 by 3, 12 channels) layers, with the later ones receiving all of the outputs of the previous ones as their input. The transition layers consist of a convolutional layer (1 by 1, same number of channels out as coming in) followed by a 2 by 2 average pooling. Dropout and batch normalisation was used when training using the softmax tops, however it was turned off when training the GPDNN model. L2 weight regularisation is used on the CNN weights. For the GPDNN model we used 80 inducing points. We use batch sizes of 64 as in [27].

Table 3: The DD-style CNN (DC) model we use on MNIST (this architecture comes from Papernot et al. [49, Table I] and is also used in Carlini and Wagner [10]).

| NN (arch. B) | GPDNN (arch C.) |
|---|---|
| Conv. (3 by 3, 32 channels) | |
| Conv. (3 by 3, 32 channels) | |
| Max pooling (2 by 2) | |
| Conv. (3 by 3, 64 channels) | |
| Conv. (3 by 3, 64 channels) | |
| Max pooling (2 by 2) | |
| FC (to 200 units) | |
| FC (to 200 units) | FC (to 50 units) |
| FC (to 10 units) | GP (RBF & WN) |
| Softmax | Robustmax |

Table 4: The DenseNet based architectures we used on CIFAR-10

| NN (arch. A) | GPDNN model (arch. C) |
|---|---|
| | Conv. (3 by 3, 24 channels) |
| | DenseNet block |
| | DenseNet transition section |
| | DenseNet block |
| | DenseNet transition section |
| | DenseNet block |
| | Average pooling (8 by 8, padding VALID) |
| FC (to 10 units) | FC (to 25 units) |
| Softmax | GP (RBF & WN) or FC (to 10 units early on in training) |
| | Robustmax (or softmax when using FC early on in training) |

When training the GPDNN model we use the ADAM optimiser, as we believe the different components may have quite different learning characteristics and so benefit from different learning rates. When using a softmax top we train via SGD as in [27]. We used a similar learning rate decay regime as [27] when using the SGD optimiser, and when training on all of the training data after 300 epochs we reach around the 7% error reported on that paper (note we do not perform any dataset augmentation).

### B.3.2 DD-style CNN (DC) architecture

When evaluating adversarial robustness of the GPDNN architecture compared to more regular NNs on the CIFAR-10 dataset we use as the base architecture the one from Papernot et al. [49, Table I] (also used in Carlini and Wagner [10]). We again call this the DD-style CNN (DC). Its structure is shown in Table 5. As well as being easier to compare results to those in [49, 10] it is also a lot shallower than the DenseNets described above enabling faster experimentation.

We add one layer of dropout after the first fully connected layer with a dropout rate of 0.5 during training. For the GPDNN we use 100 inducing points. As the accuracy of these models is a lot lower than the others we have been testing we believe that it may be better to learn the robustmax's $\beta$ value so we additionally try this. We train both models using the ADAM optimiser for 100 epochs with batch sizes of 128. The GPDNN is able to be trained up from scratch, without the bootstrapping technique done to allow it to train on top of the DenseNet. The CIFAR-10 dataset was normalised between 0 and 1 for training but we do not perform any data augmentation.

The results for the networks on the CIFAR-10 test set is shown in Table 6. The GPDNNs like in other experiments have slightly smaller error rates than the NN model. The LL for the GPDNN is worse than the NN if you fix $\beta$. This was from being penalised on test examples that it overconfidently got wrong. However, by learning $\beta$ the model has an additional place to learn its own uncertainty and the log likelihood of the GPDNN with learnt $\beta$ is the best.

Table 5: The DD-style CNN (DC) model we use on CIFAR-10 (this architecture comes from Papernot et al. [49, Table I] and is also used in Carlini and Wagner [10]) .

| NN (arch. B) | GPDNN (arch. C) |
|---|---|
| Conv. (3 by 3, 64 channels) | |
| Conv. (3 by 3, 64 channels) | |
| Max pooling (2 by 2) | |
| Conv. (3 by 3, 128 channels) | |
| Conv. (3 by 3, 128 channels) | |
| Max pooling (2 by 2) | |
| FC (to 256 units) | |
| FC (to 256 units) | FC (to 50 units) |
| FC (to 10 units) | GP (RBF + WN) |
| Softmax | Robustmax |

When learning the $\beta$ for the GPDNN values of around 0.04 were found at the end of the optimisation. This would limit the log likelihood for a point to a minimum of -5.4. and the maximum to be -0.04.

Table 6: The results of the NN (arch B DC) and GPDNN (arch C DC) on CIFAR-10 test set after being trained for 100 epochs on the whole training set. We show error rates and log likelihoods (LL). We compare two GPDNN models: one has the robustmax's $\beta = 1 \times 10^{-3}$, the other learns this parameter.

| NN | | GPDNN ($\beta$ 1E-3) | | GPDNN (learnt $\beta$) | |
|---|---|---|---|---|---|
| Error | LL | Error | LL | Error | LL |
| 0.231 | -1.235 | 0.223 | -1.612 | 0.225 | -1.09 |

# C  Adversarial examples

## C.1  FGSM

### C.1.1  Why might the FGSM sometimes fail with classifiers of greater nonlinearity?

We saw in the main text how the adversarial attack does not perform well when operating on the GPDNN network on MNIST. We can consider the FGSM attack in a simple 2D case as is done in Figure 9 to gain intuition for why this may be the case on a simple dataset such as MNIST in which it is likely that the classes at least in the final layer are linearly separable.

The FGSM is not an iterative attack. This means that the direction that it moves off in to reduce the likelihood of its current class may not be an optimal direction as it moves (unless it is a linear classifier). This means that with the RBF kernel, which may produce classifications that look like Figure 9b, although the attacker may move away from the high probability area and the likelihood will go down, it does not necessarily move towards the other class, and so the adversarial point is still classified correctly.

Having said this, it is hard to know how much this applies to deeper networks as Liu et al. [38, Figure 3] shows these classifiers on big datasets can have highly non linear decision boundaries. This could explain why we do not find exactly the same results when applying the FGSM method to the models trained on CIFAR-10.

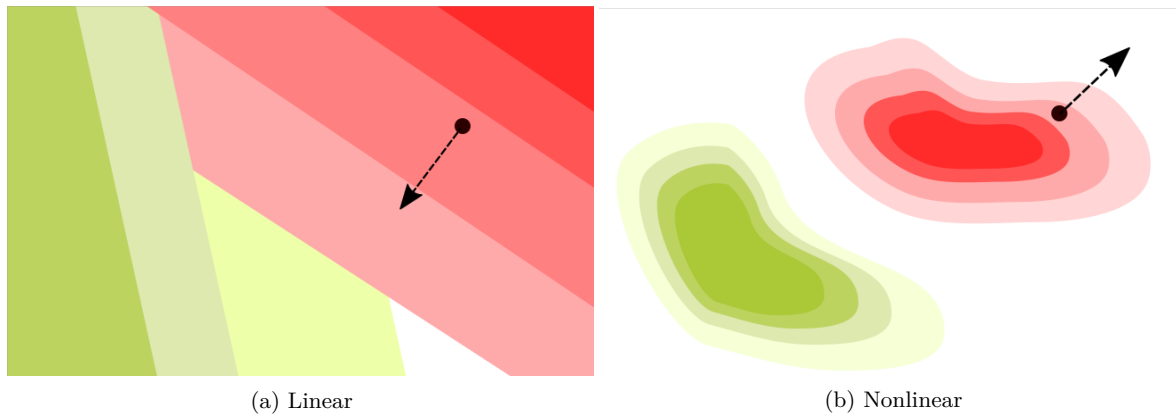

(a) Linear

(b) Nonlinear

Figure 9: Simple 2D classification problems and what one could expect a nonlinear and linear classifier to look like, on a simple easily separable dataset. On the diagram we have placed arrows showing the directions the FGSM might move in (note that due to the sign(.) function, they will not move exactly with gradient).

### C.1.2 MNIST results – applying both models to the adversarial examples generated attacking the GPDNN model

In the main text we showed the GPDNN and NN (SC) models' performances when applied to the adversarial examples generated to attack the NN (SC) model. In Figures 10, 11 and 12 we show the same metrics but when applied to the adversarial examples generated attacking the GPDNN model.

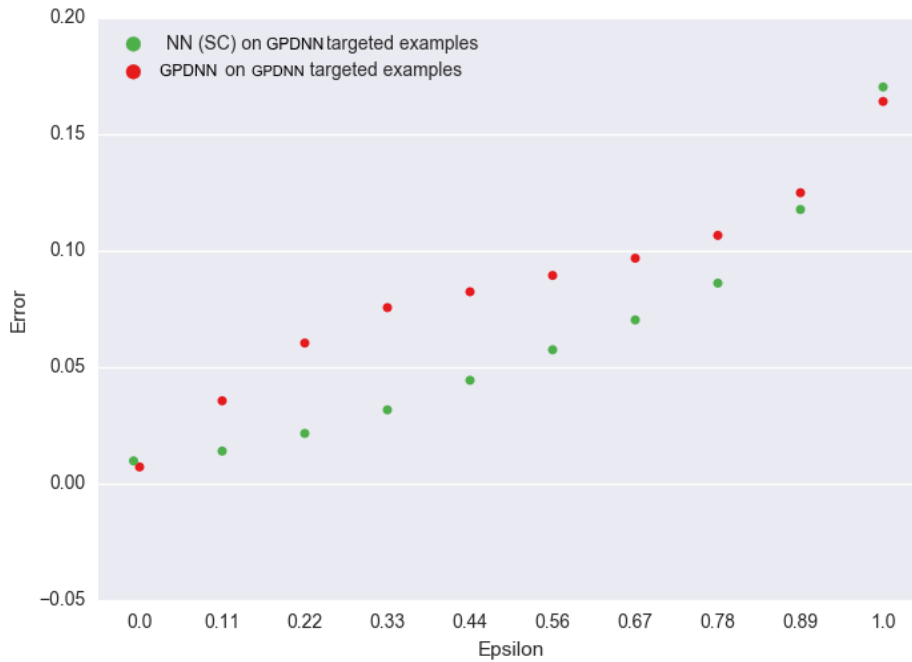

Figure 10: Error rates when applying the models to the FGSM adversarial examples generated by attacking the GPDNN model

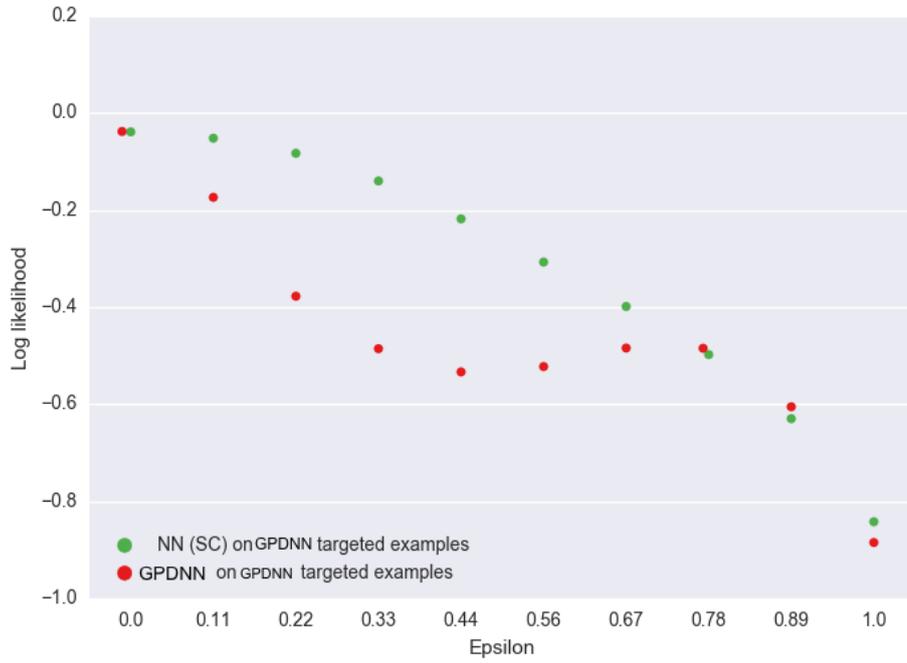

Figure 11: Log likelihoods when applying the models to the FGSM adversarial examples generated by attacking the GPDNN model

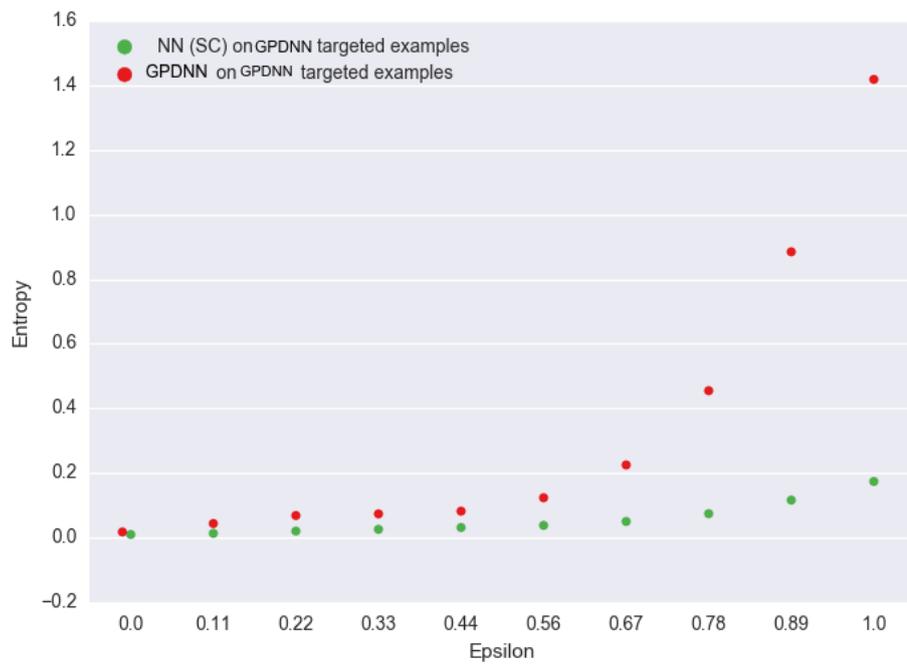

Figure 12: Entropies of predictions when applying the models to the FGSM adversarial examples generated by attacking the GPDNN model

### C.1.3 CIFAR-10 results

In this section we apply the FGSM attack to the networks trained on the CIFAR-10 dataset (those based on the DD-style (DC) network described in Section B.3.2). Note that as these base networks have a non negligible number of misclassifications on a non perturbed dataset (see §B.3.2 and in particular Table 6) we exclude from our analysis any test set examples that start off being misclassified by any of the models under consideration.

In this section we show:

- One does not have to go as far to find adversarial examples compared to when working with MNIST, perhaps as it is a more complicated space.
- That learning the robustmax's $\beta$ improves performance for the GPDNN.
- That similar to the FGSM results on MNIST above, the GPDNN generally does better with lower drop offs in error rate and log likelihood and higher predictive entropies.
- That differently to before the transfer of this attack is low (ie examples from one model attacking shown to another do not work well).
- The attack is more effective at attacking the GPDNN than it was on MNIST.

We begin by looking at the attack's effectiveness when applied to a GPDNN ($\beta = 10^{-3}$), GPDNN ($\beta$ learnt see §B.3.2) and an NN. This is shown in Figure 13. Here we see that the two GPDNN's do better in terms of smaller increase in error rate and smaller drops in log likelihood, with the GPDNN where $\beta$ is learnt doing the best. Note that the log likelihood is lower bounded for the GPDNN due to the use of the robustmax as discussed in §B.1.

No longer do we see the FGSM attack failing to do well on the GPDNN (see §C.1.1). This could be because we are in a lot more complicated space. The classifiers are getting much lower accuracies in general so that we perhaps do not have nice distinct class clusters in features being fed into the GP. This may mean that the distinct cluster hypothesis of Figure 9 breaks down.

Figure 13 makes it clear that the FGSM's step length, $\epsilon$, does not have to be as large as it was on MNIST to have as great an effect. Therefore, in Figure 14 we look further at the results for smaller $\epsilon$s. We only compare the GPDNN (learnt $\beta$) with the NN. Note that we do not use a linear scale on the x axis and we pick out the first point to be equivalent to incrementing/decrementing each subpixel by approximately one[4]. We notice that both models are very fragile: one does not need to go far to fool the model, although the GPDNN does better.

---

[4]Note that the images are normalised between 0 and 1 and have 256 levels, with $1/256 = 3.9 \times 10^{-3}$. Although the input going into the NN is not set to discrete pixel values this gives an indication on the magnitude of the changes this attack represents

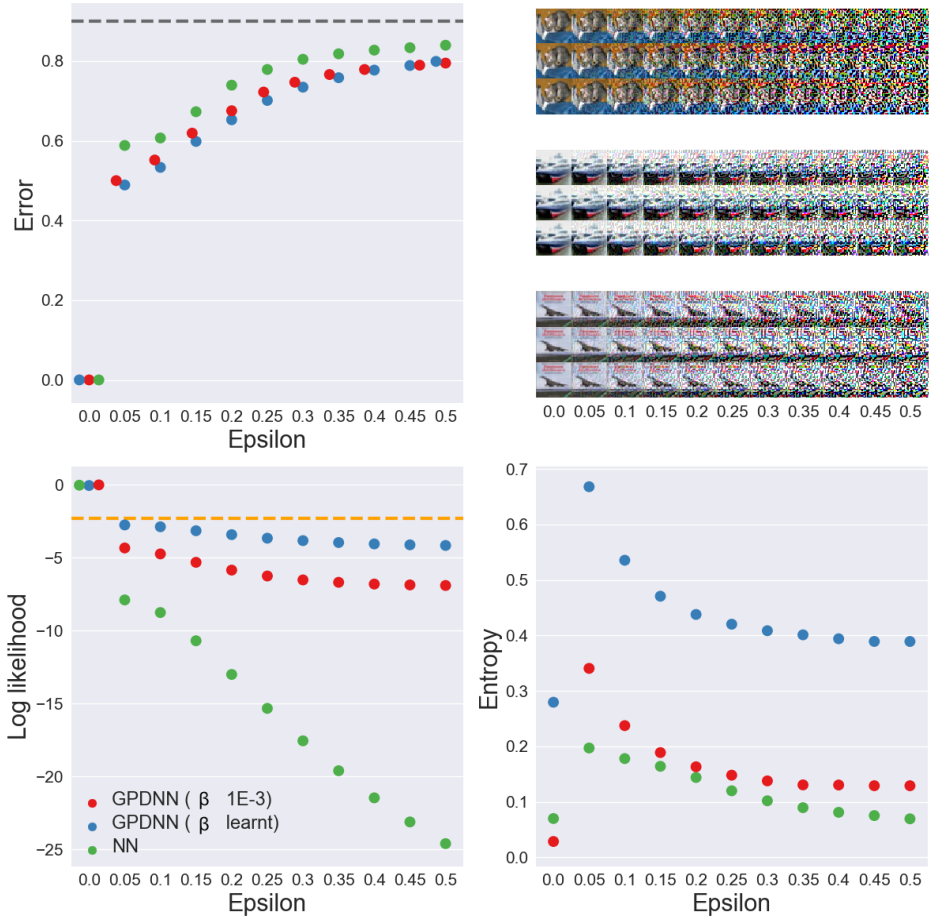

Figure 13: How the error rate, log likelihood and predictive entropies vary for the models when evaluated on adversarial examples attacking the model they are later evaluated on. The upper right plot shows the examples generated on three members of the CIFAR test set. For each individual image the rows show the example attacking the NN, the example attacking the GPDNN with the learnt $\beta$ value and the example attacking the GPDNN with $\beta$ set at $1 \times 10^{-3}$. $\beta$ in the legend refers to the $\beta$ of the robustmax, whereas the 'epsilon' in x axis refers to the $\epsilon$ or distance of the attack. The images are normalised between 0 and 1 so the distance of the attack should be judged to this. The orange dotted line shows the log likelihood of a uniform classifier always assigning 0.1 probability to each class.

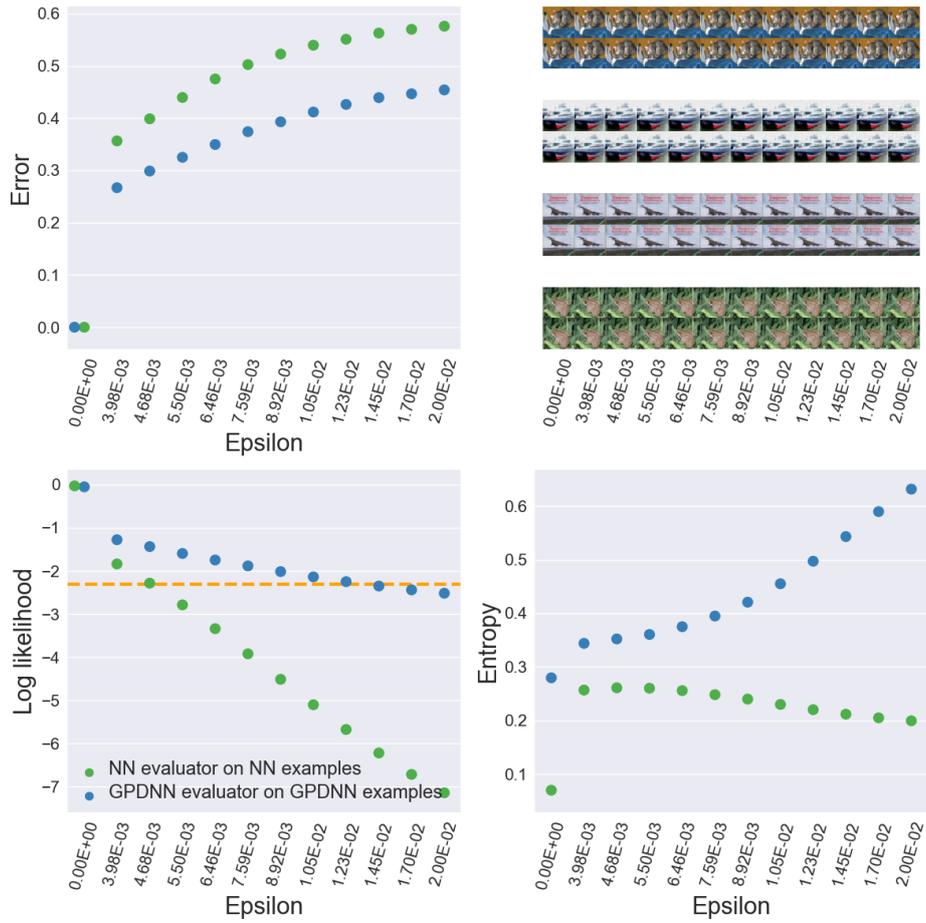

Figure 14: How the error rate, log likelihood and predictive entropies vary for the models when evaluated on adversarial examples attacking the model they are later evaluated on. The upper right plot shows examples generated on four members of the CIFAR-10 test set. For each individual image the two rows show first the example attacking the NN, then the example attacking the GPDNN with the learnt $\beta$ value, the columns correspond to the same $\epsilon$ used in the other subplots. The orange dotted line shows the log likelihood of a uniform classifier always assigning 0.1 probability to each class.

**Comparing the models on the examples attacking the other model**  Finally, we look at applying the examples generated by each model on the other model. We start off by applying both models to the examples attacking the NN model in Figure 15. We then do the converse thing in Figure 16 where we look at each model evaluating the examples generated attacking the GPDNN. We use GPDNNs with the learnt $\beta$.

From this we see at the examples transfer poorly. This in contrast to what we were seeing in the MNIST experiments. Perhaps this is because we're using deeper architectures on a more complicated space so the representations learnt by the two models may differ more. Also the relative step sizes, $\epsilon$ we are looking at are smaller.

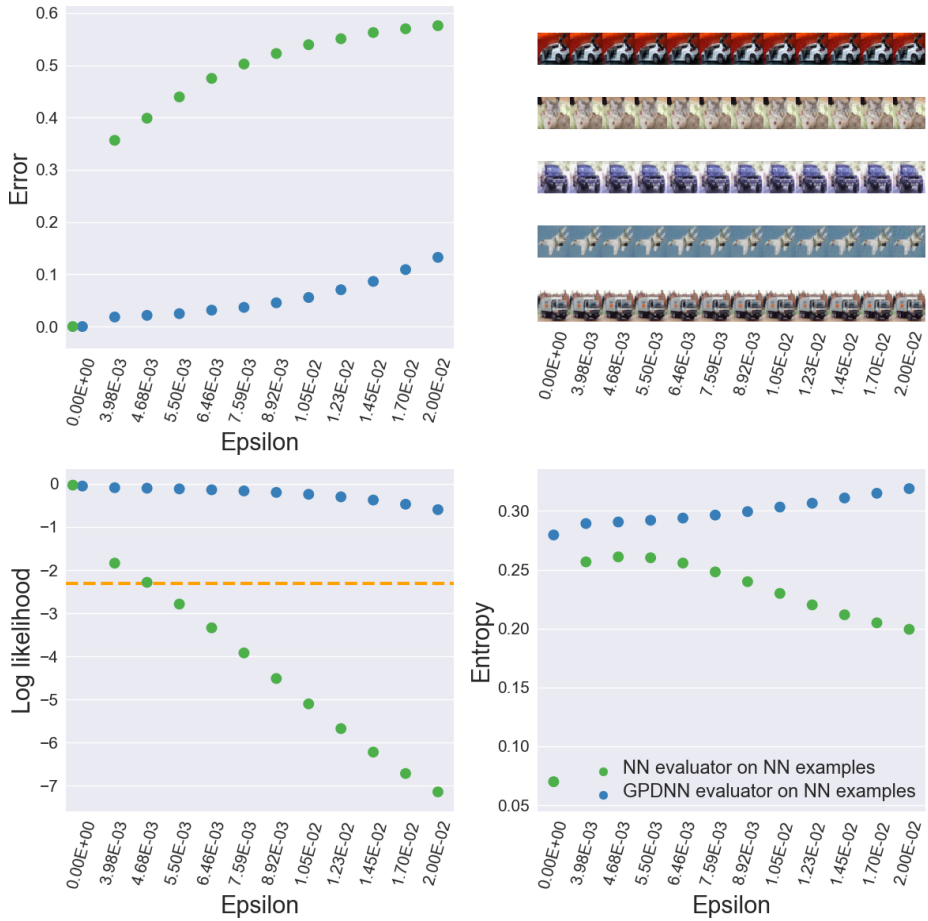

Figure 15: How the error rate, log likelihood and predictive entropies vary for the models when evaluated on adversarial examples attacking the NN model. The upper right plot shows the examples generated on five members of the CIFAR test set. The orange dotted line shows the log likelihood of a uniform classifier always assigning 0.1 probability to each class.

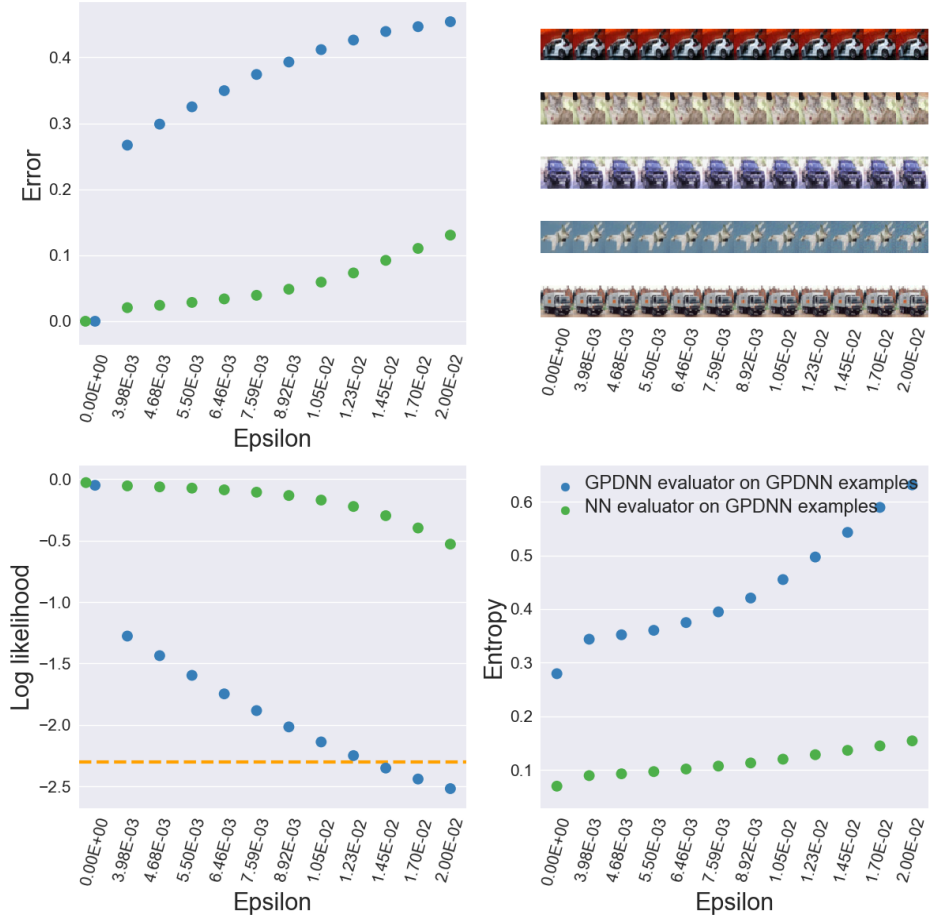

Figure 16: How the error rate, log likelihood and predictive entropies vary for the models when evaluated on adversarial examples attacking the GPDNN model. The upper right plot shows the examples generated on five members of the CIFAR-10 test set. The orange dotted line shows the log likelihood of a uniform classifier always assigning 0.1 probability to each class.

## C.2 L2 optimization attack of Carlini and Wagner [10]

In this section we first give details of how this attack was applied to the GPDNN (where we do not have obvious pre-softmax values to attack). We then give further details of the results of the attack on the two models by:

1. looking at the error rates of the successful attack examples on one model applied to other models.

2. showing the distribution of the distances that the attacker went on one model compared to another on the same image.

3. showing examples of the adversarial examples suggested and some of their predictions.

We first do this for the MNIST dataset (models based on the same small CNN architecture used for the FGSM attack). We then repeat the same analysis for the DD-style CNN (DC) models trained on the CIFAR-10 dataset.

### C.2.1 How we applied this attack to the GPDNN

The L2 optimization attack of Carlini and Wagner [10] works on pre-softmax values. When applying this attack to the hybrid model we have a problem as we do not have pre-softmax values that we can use. The latent variables that are fed into the robustmax for each class come in the form of a probability distribution. We therefore take the probabilities from the robustmax and calculate what would have been the pre-softmax difference had a softmax been used. This should give sensible values to use with the search setup of Carlini and Wagner [10].

Note that in both cases of the attack Carlini and Wagner [10] actually optimise $\mu$ in a transformed space so that they do not go out of the image bounds. We do the same.

### C.2.2 MNIST results

**Transferability** As discussed in the main text, 381 out of the 1000 attacks on the GPDNN model failed in the sense the algorithm did not find any suggestion. None of the L2 optimisation attacks failed in this manner when attacking the regular NNs. This is promising for the GPDNN model, although we warn that we used only the default parameters for this attack[5] and did not spend time tuning the algorithm, so perhaps this attack could be made more effective. However, we hope that as we converted the robustmax values to what they would have been pre-softmax (up to a constant) if a softmax top had been used, the default algorithm settings were sensible ones to try.

Table 7 shows the error rates and log likelihoods as we evaluate our models (along the columns) across the adversarial examples generated by the L2 attacks on the different models (along the rows), as well as plain MNIST at the top. We do not include the failed attacks from the GPDNN model in this table. The transfer of this attack is low but Liu et al. [38, §3.1] found comparable results with a similar attack on the ImageNet database. Their solution to generate examples that transferred better was to run the ADAM optimiser for longer so that the perturbation grew larger (they also did not have a distance penalising term on the perturbation).

Table 7: How the L2 optimisation adversarial attack examples transfer between the different models,† note that this result is on the full MNIST test set rather than a subset of 1000, and no adversarial perturbations have been made. Also note that the GPDNN results are only on the attacks that succeeded.

| Model targeted | NN (SC) Error | LL | GPDNN (SC) Error | LL | NN (DC) Error | LL |
|---|---|---|---|---|---|---|
| Plain MNIST † | 0.0100 | -0.039 | 0.007 | -0.038 | 0.004 | -0.021 |
| NN (SC, arch. A) | 1.000 | -0.718 | 0.009 | -0.056 | 0.003 | -0.014 |
| GPDNN (SC, arch C) | 0.031 | -0.094 | 1.000 | -0.828 | 0.011 | -0.056 |
| NN (DC, arch B) | 0.014 | -0.047 | 0.013 | -0.067 | 1.000 | -0.725 |

**Differences in distances** Figures 17-19 show a histogram of the differences in perturbations suggested by the algorithm when applied to the different models. Note these distances can be compared to the range of the pixel values which are between -1 and 1. The clump at around 25 on the GPDNN plots represents the failed attacks where the algorithm arbitrarily returns images of all zeros.

Figure 18 shows a histogram of the differences in Euclidean distance that the GPDNN (SC) model's adversarial perturbations are compared to the NN (DC). One can see that similar to comparing against the regular NN (SC) the perturbations needed are greater in distance. Figure 19 is the same plot except for the NN (DC) model against the NN (SC). As one can see most of the differences are greater than zero, meaning that the examples found are perturbed further for the NN (DC). It suggests that the deeper and better performing NN (DC) model is more adversarially robust than the shallower model but not as robust as the GPDNN model.

---

[5]described in Carlini and Wagner [10]'s code available at `https://github.com/carlini/nn_robust_attacks`

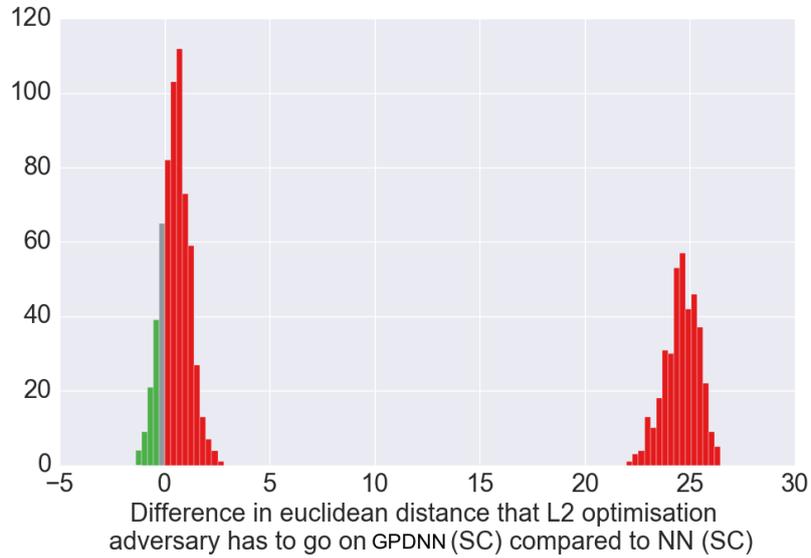

Figure 17: Differences in magnitudes of perturbation needed between GPDNN network and NN (SC). The red region (above 0) shows areas for which the attacker has had to perturb the GPDNN adversarial example by a greater distance than the NN (SC) adversarial for the same image.

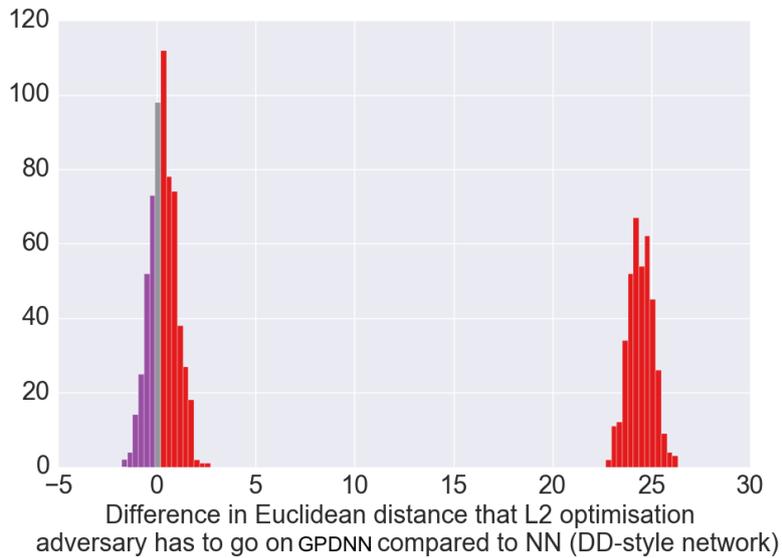

Figure 18: Differences in magnitudes of perturbation needed between GPDNN (SC) network and NN (DC) network. The red region (above 0) shows areas for which the attacker has had to perturb the GPDNN adversarial example by a greater distance than the NN (DC) adversarial for the same image.

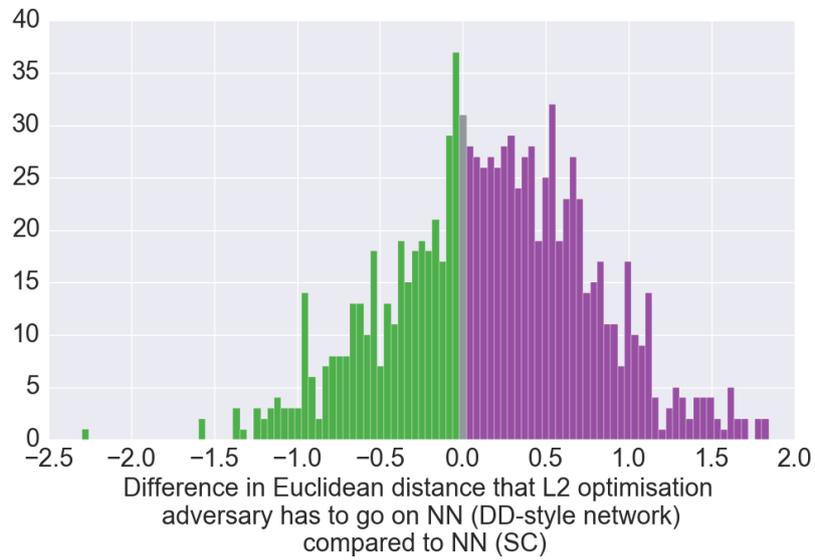

Figure 19: Differences in magnitudes of perturbation needed between NN (SC) network and NN (DC). The purple region (above 0) shows areas for which the attacker has had to perturb the NN (DC) adversarial example by a greater distance than the NN (SC) adversarial for the same image.

**Adversarial examples** In Figure 20 we show the adversarial samples generated from the L2 optimisation algorithm when attacking the NN (SC) and the GPDNN models. Failed attacks take the form of a grey square.

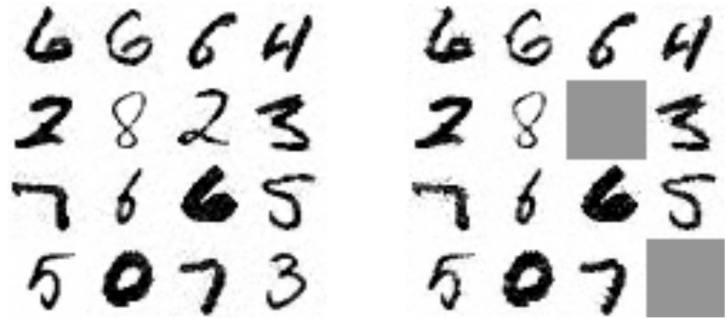

Figure 20: Carlini examples generated attacking the NN (SC) model left and the GPDNN model right

**Prediction examples** Figures 21, 22, and 23 show the adversarial examples generated when attacking the GPDNN (SC) model and NN (SC) model. We also show the model predictions on the respective images. The left hand column is for plain MNIST, the middle for the samples generated attacking the GPDNN model (SC arch. C) and the right the examples generated when attacking the NN (SC arch. A). The first row shows the images the model is classifying. The second row shows the predictions made by the GPDNN model for this image, and the third and final row shows the predictions made by the NN (SC) model.

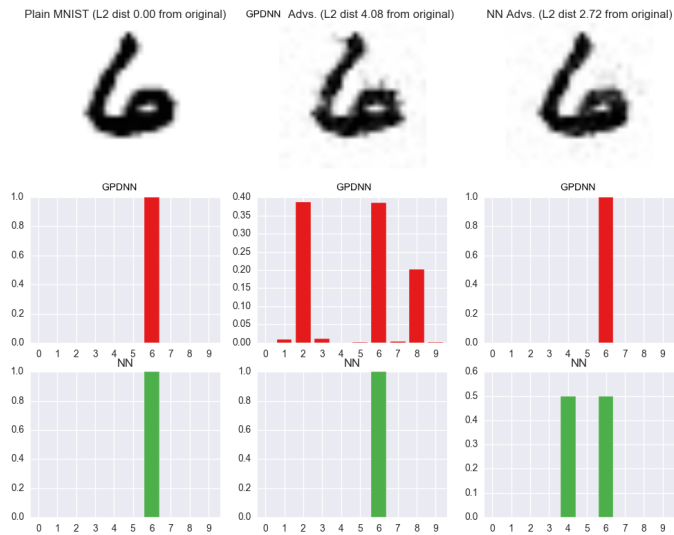

Figure 21: Adversarial example generated on MNIST for the NN (SC) and GPDNN networks and how the respective models predict.

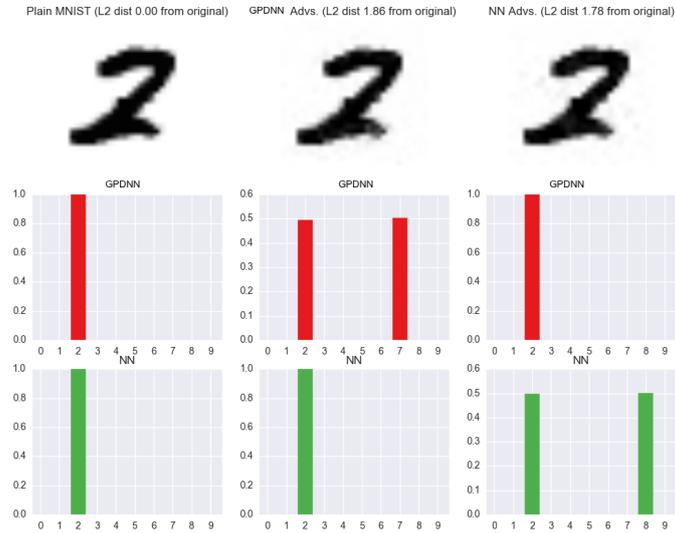

Figure 22: Adversarial example generated on MNIST for the NN (SC) and GPDNN networks and how the respective models predict.

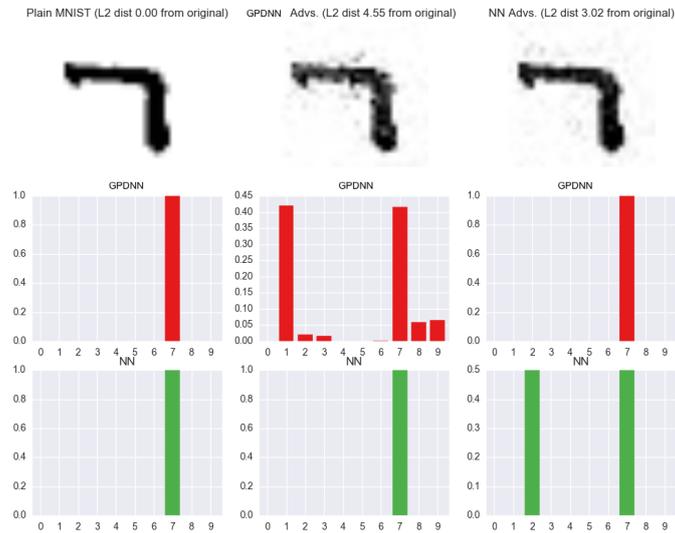

Figure 23: Adversarial example generated on MNIST for the NN (SC) and GPDNN networks and how the respective models predict.

### C.2.3 CIFAR-10 Results

We can do the same analysis using the CIFAR-10 dataset. This is briefly described in this section. We use the same DD-style CNN networks used for the FGSM attack on CIFAR-10. In effect this is comparing an architecture B network with an architecture C one. We allow the robustmax's $\beta$ to be learnt again.

To evaluate we pick 1000 random images that are classified correctly by both classifiers. We then learn the Carlini L2 algorithm the same way it was run for MNIST.

**Transferability** On CIFAR-10 the attack fails 207 times when attacking the GPDNN but succeeded all the time when attacking the more regular NN. The error rates and the log likelihoods when the two models are used to evaluate examples generated on each model is shown in Table 8. Like for MNIST this attack has poor transferability between the models. The values for the original examples are not shown as they were picked to have zero error rates before being attacked. The log likelihoods were -0.02 for the NN and -0.05 for the GPDNN.

Table 8: Transferability of the adversarial examples generated by the two networks when evaluated on the two networks. The rows correspond to the model that the attacker was targeting whereas the columns describe the model used to do the evaluations. When testing on the examples generated through attacking the GPDNN we remove the attacks that failed.

| Model targeted | NN (DC) Error | NN (DC) LL | GPDNN (DC) Error | GPDNN (DC) LL |
|---|---|---|---|---|
| NN (DC, arch B) | 1 | -0.788 | 0.012 | -0.08 |
| GPDNN (DC, arch C) | 0.016 | -0.066 | 1 | -0.935 |

**Differences in distances** Figure 24 shows the pairwise differences in distance found by the attacker working on the GPDNN (DC) model compared to when attacking the NN (DC) model. On average (excluding the failed attacks) the attacker went a distance of 0.071 further when attacking the GPDNN model. The images were normalised between 0 and 1 and are three channel so this corresponds to increasing one pixel by about 10 integer values over all subpixels or to changing around 109 pixels by one integer level.

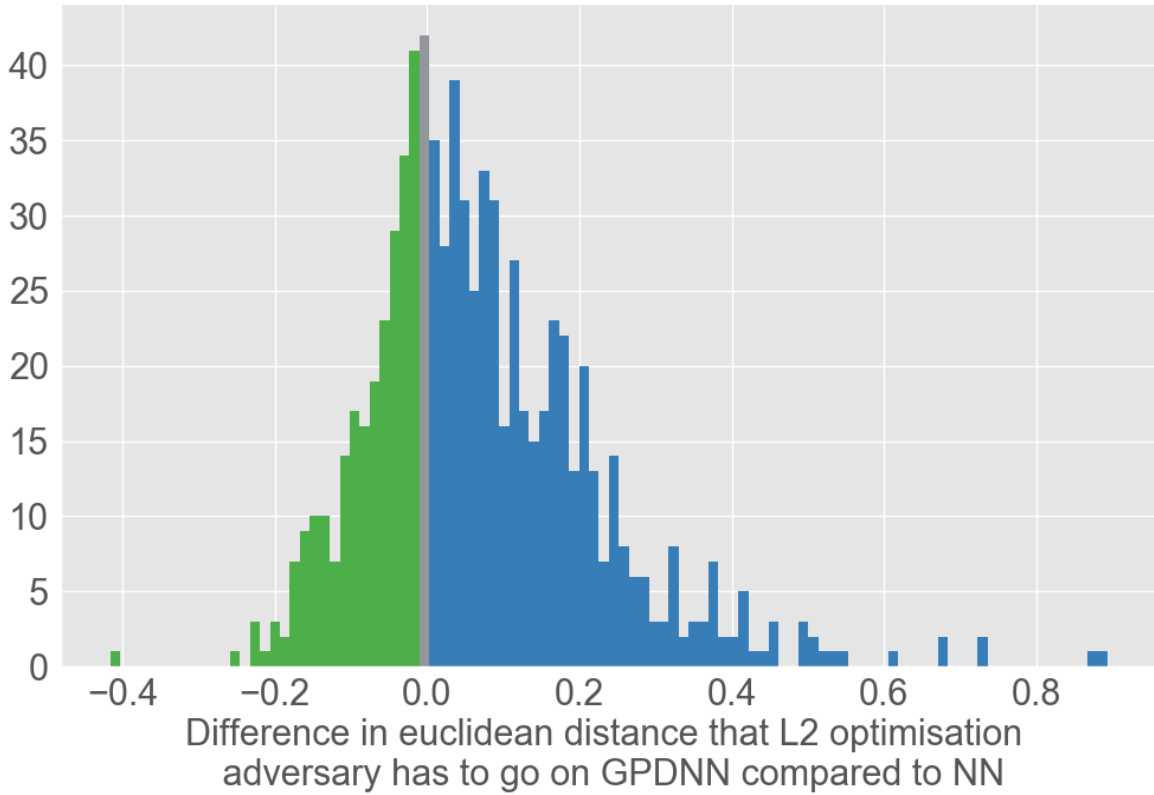

Figure 24: Differences in magnitudes of perturbation needed between GPDNN (DC) network and NN (DC). The blue region (above 0) shows areas for which the attacker has had to perturb the GPDNN adversarial example by a greater distance than the NN adversarial example for the same image. We have removed the 207 attacks that failed on the GPDNN from this image.

**Adversarial examples** We show some examples of the attack results in Figure 25.

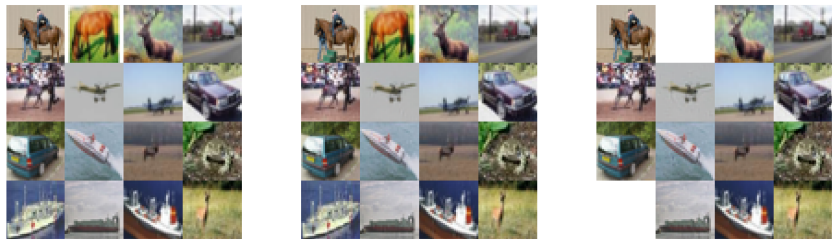

Figure 25: On the left we have original examples from the CIFAR test set. In the middle we have examples attacking the NN and on the right we have the examples generated by the L2 optimization attack on the GPDNN. Failed attacks are shown in white.

Below we also magnify some of the examples in Figures 26 and 27. In Figure 26 one can see some changes to the sky around the airplane when looking carefully. With Figure 27 it is hard to notice any difference visually.

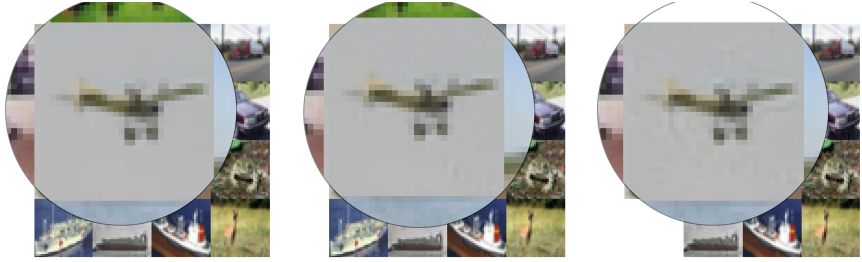

Figure 26: Magnified example of the attacks on an airplane. The NN attack has gone a distance of 0.19 from the original whereas the attack on the GPDNN has gone a distance of 0.32. Original image is on the left, example attacking the NN in the middle and the example attacking the GPDNN on the right.

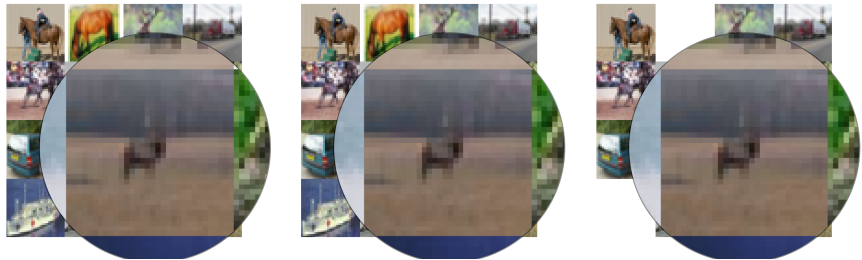

Figure 27: Magnified example of the attacks on a deer. The NN attack has gone a distance of 0.04 from the original whereas the attack on the GPDNN has gone a distance of 0.08. Original image is on the left, example attacking the NN in the middle and the example attacking the GPDNN on the right.

**Example predictions** As for MNIST we show some example predictions made on these adversarial examples.

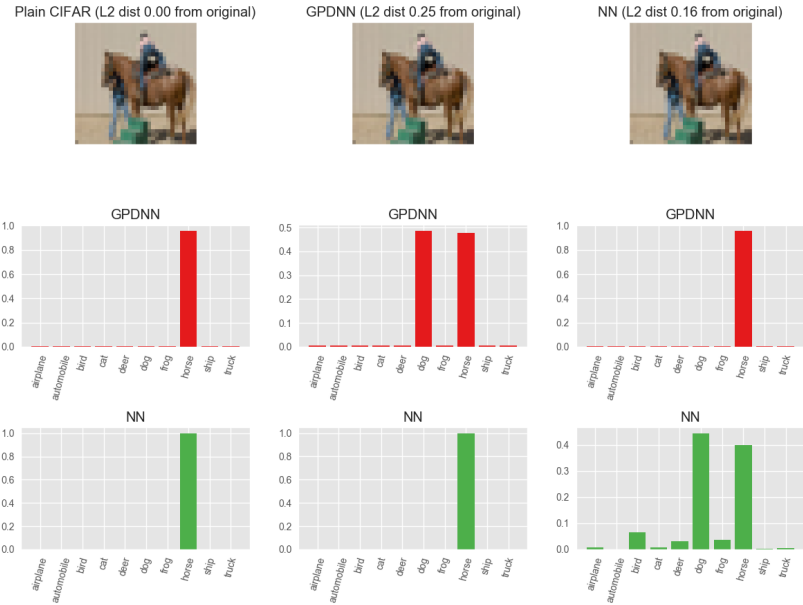

Figure 28: Example of the two model predictions on the original image from the CIFAR-10 test set, and the adversarial examples attacking the GPDNN and NN. The columns show the original image, the GPDNN attacking image and the NN attacking image respectively. The rows show what the image looks like their predictions from the GPDNN and their predictions from the NN.

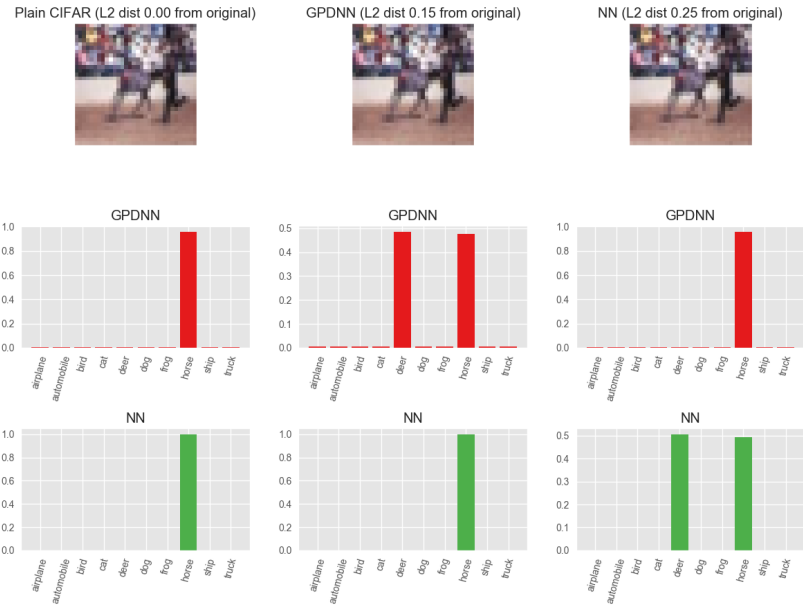

Figure 29: Example of the two model predictions on the original image from the CIFAR-10 test set, and the adversarial examples attacking the GPDNN and NN. The columns show the original image, the GPDNN attacking image and the NN attacking image respectively. The rows show what the image looks like their predictions from the GPDNN and their predictions from the NN.

# D    Transfer testing

Here we give a little more detail on the other datasets we used in for transfer testing and how they were preprocessed.

- ANOMNIST is a dataset that we created. It consists of 900 digits (90 of each type) drawn by one of the authors of this paper. The digits were drawn with a little padding in a similar style to MNIST.
- The second new dataset we consider is the Semeion dataset [6] [1, 8]. This consists of 1593 handwritten digits from 80 writers. The original images are 16 by 16 pixels so we scale up to 28 by 28 to get the same size as MNIST.
- The final dataset we consider is the Street View House Numbers dataset [45]. This consists of 26032 images (as we only consider the test set). We scaled these images down in size and converted to greyscale.

---

[6]available from `http://archive.ics.uci.edu/ml/datasets/semeion+handwritten+digit`